\documentclass[11pt]{article}
\usepackage{eamt26}
\usepackage{times}
\usepackage{url}
\usepackage{latexsym}
\usepackage[small,bf]{caption} 
\usepackage{multirow}
\usepackage{booktabs}
\usepackage{arydshln}
\usepackage{float}
\usepackage{amsmath,amsfonts,amssymb}
\usepackage{graphicx}
\usepackage{xspace}
\usepackage[T1]{fontenc}

\newcommand{\QwenLargeInstruct}{\text{Qwen3-30B-A3B-Instruct-2507}\xspace}
\newcommand{\QwenLargeThink}{\text{Qwen3-30B-A3B-Thinking-2507}\xspace}
\newcommand{\QwenSmallInstruct}{\text{Qwen3-4B-Instruct-2507}\xspace}
\newcommand{\QwenSmallThink}{\text{Qwen3-4B-Thinking-2507}\xspace}
\newcommand{\Llama}{\text{Llama-3.2-3B-Instruct}\xspace}
\newcommand{\Gemma}{\text{gemma-3-27b-it}\xspace}
\newcommand{\QwenOld}{\text{Qwen2.5-32B-Instruct}\xspace}
\newcommand{\DeepseekQwen}{\text{DeepSeek-R1-Distill-Qwen-32B}\xspace}
\newcommand{\Aya}{\text{aya-expanse-32b}\xspace}
\newcommand{\Tower}{\text{Tower-Plus-72B}\xspace}
\newcommand{\TFiveGemma}{\text{t5gemma-xl-xl-prefixlm-it}\xspace}
\newcommand{\DeepseekChat}{\text{DeepSeek-V3.2-Exp-671B-chat}\xspace}
\newcommand{\DeepseekReasoner}{\text{DeepSeek-V3.2-Exp-671B-reasoner}\xspace}
\newcommand{\NllbSmall}{\text{nllb-200-3.3B}\xspace}
\newcommand{\NllbBig}{\text{nllb-moe-54b}\xspace}
\newcommand{\GoogleTranslate}{\text{Google Translate}\xspace}

\title{Why do Large Language Models Fail in Low-resource Translation? Unraveling the Token Dynamics of Large Language Models for Machine Translation}

\author{
Shenbin Qian \and Yves Scherrer \\
Language Technology Group, Department of Informatics \\
University of Oslo, Norway \\
\texttt{\{shenbinq, yves.scherrer\}@ifi.uio.no}
}

\date{}

\begin{document}
\maketitle
\begin{abstract}
Large Language Models (LLMs) have recently demonstrated strong performance in machine translation (MT). However, most prior work focuses on improving or benchmarking translation quality, offering limited insight into when and why LLM-based translation fails. In this work, we systematically analyze failure modes of LLMs in MT by evaluating 15 models, including four reasoning LLMs, across 22 language pairs (LPs) with varying resource levels. We find that non-English-centric LPs consistently yield lower COMET scores than English-centric pairs. To investigate the underlying causes, we introduce \textbf{Token Activation Rate (TAR)}, a metric that captures how effectively a model utilizes language-specific tokens in its vocabulary during generation. We validate TAR as a proxy for language representation using models with known language distributions in the training data, and show that lower TAR is strongly associated with poorer translation performance. Furthermore, reasoning LLMs tend to generate more tokens when translating into low-TAR languages, suggesting a compensatory mechanism, although its impact on translation quality varies across models. Overall, our findings emphasize the importance of token-level dynamics in understanding MT performance of LLMs.
\end{abstract}

\section{Introduction}

Large Language Models (LLMs) have achieved significant advancements across various subfields of Natural Language Processing (NLP), including sentiment analysis, text summarization, and machine translation (MT) \cite{zhang-etal-2024-sentiment,Pu2023-eo,ZhangBiao2023}. More recently, LLMs trained via Reinforcement Learning with Verifiable Rewards (RLVR) \cite{Lambert2024-vb} have demonstrated reasoning capabilities that extend beyond language tasks to include coding and mathematical problem-solving \cite{OpenAI2024-xa,Guo2025-mj,ahn-etal-2024-large,Jiang2025}.

Alongside these developments, numerous benchmarks have emerged to evaluate the state-of-the-art capabilities of LLMs on specific tasks \cite{Wang2019etal,hendrycks2021measuring,OpenAI2024Bench,Phan2025-io,yue-etal-2025-mmmu,romanou2025include,huang-etal-2025-benchmax}. However, most benchmarks aim to assess how well LLMs perform on tasks with definitive correct answers, typically through multiple-choice formats or comparison with human-prepared references, but not on open-ended multilingual generation tasks like translation. Although MT evaluation datasets such as FLORES \cite{guzman-etal-2019-flores} or test sets from the Conference on Machine Translation (WMT\footnote{\url{https://www2.statmt.org/}}) can be leveraged for evaluating LLMs' translation abilities, relatively little work has investigated why LLMs fail on certain translation tasks, particularly in low-resource and non-English-centric settings.

To address this gap, we perform a large-scale empirical analysis of LLM-based translation, focusing on how performance varies across language pairs (LPs) with different resource availabilities. We observe that non-English-centric and lower-resource LPs consistently yield lower COMET \cite{rei-etal-2020-comet,stewart-etal-2020-comet,rei-etal-2022-comet,rei-etal-2022-cometkiwi} and BLEU \cite{papineni-etal-2002-bleu} scores. We hypothesize that low token activation for these languages contributes to these failures, and that reasoning models may partially compensate by generating more tokens at inference time. Our contributions are as follows:

\begin{itemize}
    \item We evaluate 15 models across 22 LPs and show that non-English-centric LPs exhibit significantly lower COMET scores compared to English-centric pairs.
    \item We propose \textbf{Token Activation Rate (TAR)}\footnote{\url{https://github.com/shenbinqian/llm4mt}} as a metric for quantifying language representation in model vocabularies, and demonstrate its effectiveness as a proxy for language coverage. We further show that TAR and \textbf{typological distance} are strongly associated with COMET and BLEU scores.
    \item We investigate the relationships among TAR, \textbf{reasoning tokens}, and COMET and BLEU scores. Our findings suggest that low TAR of the target language is significantly correlated with the number of generated reasoning tokens, which for some LLMs is correlated with COMET or BLEU improvements.
\end{itemize}

\section{Related Work}

\paragraph{LLM Translation} The emergence of LLMs has spurred extensive research on their application to machine translation \cite{ZhangBiao2023,vilar-etal-2023-prompting,castaldo-monti-2024-prompting,he-2024-prompting}. Early work \cite{ZhangBiao2023} explored prompting strategies and showed that well-designed prompts can yield performance comparable to traditional MT systems. Subsequent studies \cite{kocmi-etal-2024-findings,Song2025-fj} highlight that LLMs consistently underperform in low-resource settings, motivating approaches such as retrieval-augmented and context-aware translation \cite{court-elsner-2024-shortcomings}. More recently, reasoning LLMs have been applied to translation tasks. Liu et al. \shortcite{Liu2025-ty} argue that these models improve contextual coherence, cultural intentionality, and self-reflection, while Ye et al. \shortcite{Ye2025-zx} show that they outperform instruction-tuned models in semantically complex domains, particularly for long-text and high-difficulty translation scenarios. Despite these advances, prior work largely focuses on improving translation quality rather than explaining the root causes of failure, particularly in low-resource settings.

\paragraph{Tokenization and Vocabulary Effects in MT} A growing body of work attributes translation failures to tokenization and vocabulary design \cite{rust-etal-2021-good,sindhujan-etal-2025-llms,Lundin2025-ta}. Multilingual models often underperform on languages that are under-represented in the shared vocabulary, while dedicated or language-specific tokenizers can mitigate this gap \cite{rust-etal-2021-good}. Tokenization inefficiency, commonly measured by high sub-word fertility, has also been shown to correlate with lower performance, especially for morphologically rich and low-resource languages \cite{Lundin2025-ta}. Several methods have been proposed to address these issues including stochastic segmentation techniques, such as BPE-dropout, vocabulary refinement approaches to remove low-utility tokens, and targeted vocabulary expansion \textit{etc} \cite{provilkov-etal-2020-bpe,chizhov-etal-2024-bpe,singh-etal-2025-information}. Overall, prior work consistently links tokenization properties such as vocabulary coverage, or token efficiency, to downstream translation performance. However, these studies primarily focus on model design and optimization, leaving open the question of how token-level dynamics within LLMs contribute to systematic failures in translation, especially low-resource settings.

\section{Experimental Setup}

We describe our datasets in Section \ref{sec:data}. Models and inference details are in Sections \ref{sec:method}  and \ref{sec:inf_details}.

\begin{table*}[ht]
\centering
\resizebox{13cm}{!}{%
\begin{tabular}{ccccc}
\toprule
Model Name                              & Architecture          & Instruction-tuned or Reasoning & Open Weights & Parameter Size         \\ \hline
\QwenLargeInstruct         & decoder-only-moe      & instruction-tuned              & yes           & 30B in total, 3B active \\
\QwenLargeThink         & decoder-only-moe      & reasoning                      & yes           & 30B in total, 3B active \\
\QwenSmallInstruct              & decoder-only-dense    & instruction-tuned              & yes           & 4B                      \\
\QwenSmallThink              & decoder-only-dense    & reasoning                      & yes           & 4B                      \\
\Llama         & decoder-only-dense    & instruction-tuned              & yes           & 3B                      \\
\Gemma                    & decoder-only-dense    & instruction-tuned              & yes           & 27B                     \\
\QwenOld                & decoder-only-dense    & instruction-tuned              & yes           & 32B                     \\
\DeepseekQwen  & decoder-only-dense  & reasoning                      & yes           & 32B                     \\
\Aya               & decoder-only-dense    & instruction-tuned              & yes           & 32B                     \\
\Tower                   & decoder-only-dense    & instruction-tuned              & yes           & 72B                     \\
\TFiveGemma         & encoder-decoder-dense & instruction-tuned              & yes           & 4B                      \\
Deepseek-V3.2-Exp                  & decoder-only-moe      & mixed                          & yes           & 671B                    \\
\NllbSmall                   & encoder-decoder-dense & neither, translation only      & yes           & 3.3B    \\
\NllbBig                   & encoder-decoder-moe   & neither, translation only      & yes           & 54B                   \\
\GoogleTranslate                         & unknown               & neither, translation only      & no            & unknown      \\      
\bottomrule
\end{tabular}%
}
\caption{Model details including names, architectures, size and either instruction-tuned or reasoning and open-weights or proprietary models.}
\label{tab:model_details}
\end{table*}

\subsection{Data} \label{sec:data}

To assess the translation capabilities of LLMs, we compiled multiple datasets covering different LPs and translation directions across resource-varying settings. Our test data comprises 10 \textbf{non-English-centric} LPs\footnote{These datasets do not involve English during the process of their construction, unlike FLORES.} and 12 \textbf{English-centric} LPs, with the latter consisting of 6 \textbf{en-XX} pairs and 6 \textbf{XX-en} pairs. These span high-, medium-, and low-resource languages, including \textbf{Arabic-Chinese (ar-zh)}, \textbf{Arabic-Hebrew (ar-he)}, \textbf{Chinese-French (zh-fr)}, \textbf{Chinese-Russian (zh-ru)}, \textbf{French-Italian (fr-it)}, \textbf{German-French (de-fr)}, \textbf{German-Italian (de-it)}, \textbf{Korean-Chinese (ko-zh)}, \textbf{Korean-French (ko-fr)}, and \textbf{Russian-French (ru-fr)} from the TED Multilingual Parallel Corpora \cite{Kulkarni2015}, the multilingual corpus from the Swiss Federal Administration (SwissAdmin) \cite{scherrer-etal-2014-swissadmin}, and the Chinese-Korean parallel corpus \cite{Park2019-vd}; as well as \textbf{English-Chinese (en-zh)}, \textbf{English-Czech (en-cs)}, \textbf{English-German (en-de)}, \textbf{English-Polish (en-pl)}, \textbf{English-Russian (en-ru)}, \textbf{English-Tamil (en-ta)}, \textbf{Chinese-English (zh-en)}, \textbf{Czech-English (cs-en)}, \textbf{German-English (de-en)}, \textbf{Khmer-English (km-en)}, \textbf{Russian-English (ru-en)}, and \textbf{Tamil-English (ta-en)} from the Quality Estimation Shared Task of the Fifth Conference on Machine Translation (WMT20) \cite{barrault-etal-2020-findings}. We randomly sampled 3,000 examples per LP from these corpora to form our test set, yielding 66,000 instances in total\footnote{We treat language pairs with different translation directions as distinct, as we used separate data instances for each direction rather than swapping source and target.} (see Appendix \ref{appendix_data}). We did not select these resources with the intention of benchmarking the latest LLMs, as they are publicly available online and may have been included in LLM training data. Rather, we use this data to investigate when and why models fail, even on potentially seen examples.

\subsection{Methodology} \label{sec:method}

\paragraph{Prompt Selection} We initially adopted the prompt template from Zhang et al. \shortcite{ZhangBiao2023} to instruct LLMs to perform translation via in-context learning in both zero-shot and few-shot settings. However, preliminary experiments revealed that some models failed to adhere to the instruction, producing verbose and noisy outputs with explanatory text rather than translations in the target language (see Appendix \ref{appendix_noise}). Such behavior interferes with reliable automatic evaluation. To deal with this issue, we designed two additional prompt templates aimed at eliciting translation-only outputs. We denote the original prompt from Zhang et al. \shortcite{ZhangBiao2023} as Prompt 0, and our proposed templates as Prompt 1 and Prompt 2 (see Appendix \ref{appendix_prompt}). These prompts are not intended to optimize translation performance, but to ensure output consistency for evaluation, which is critical for maintaining the validity of metric-based comparisons such as COMET and BLEU. We conducted experiments with all 3 prompts and assessed output noise using a rule-based detector followed by manual inspection (see Appendix \ref{appendix_verifier}). We selected outputs from Prompt 2, which consistently produced the cleanest translations, for all subsequent analyses.

\begin{figure*}[ht]
  \centering
  \includegraphics[width=0.99\textwidth]{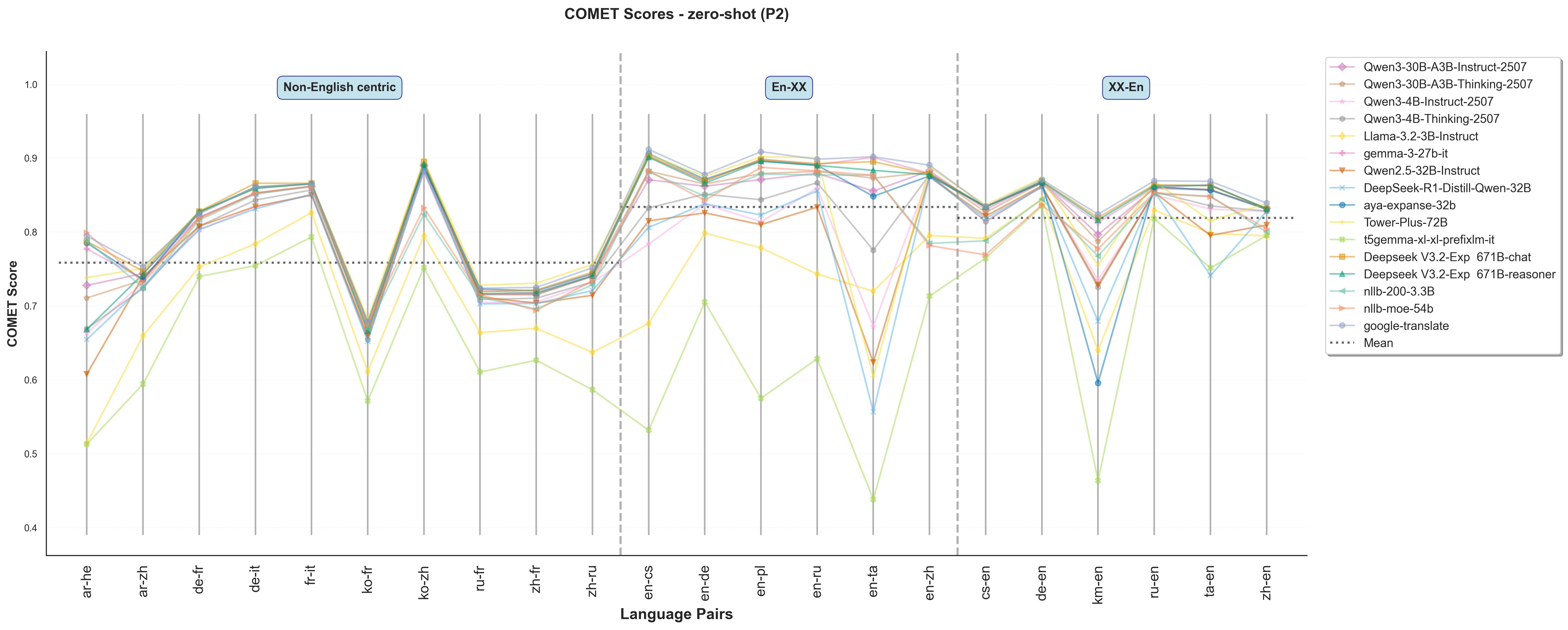}
  \caption{COMET scores of translations for 22 language pairs using Prompt 2 under zero-shot setting.}
  \label{fig.zero-shot_comet_p2}
\end{figure*}

\paragraph{Model Selection} We selected 15 models spanning a wide range of sizes, architectures, post-training methods, and levels of multilingual data coverage as shown in Table \ref{tab:model_details}. These include decoder-only instruction-tuned (IT) models from the Qwen series, such as \textbf{\QwenLargeInstruct} and \textbf{\QwenSmallInstruct}, along with their corresponding reasoning variants post-trained using RLVR: \textbf{\QwenLargeThink} and \textbf{\QwenSmallThink} \cite{QwenTeam2025}. To compare instruction-tuned and reasoning models, we also include \textbf{\QwenOld} \cite{QwenTeam2024} versus \textbf{\DeepseekQwen}, which share the same base model but differ in post-training—the latter was trained via knowledge distillation \cite{Hinton2015-ny} using DeepSeek-R1 \cite{Guo2025-mj} as a teacher model trained with RLVR. Additionally, we compare the chat mode and reasoning mode of \textbf{DeepSeek-V3.2-Exp} (\DeepseekChat and \DeepseekReasoner, respectively) \cite{deepseekai2024deepseekv32}. \textbf{\Llama} \cite{meta2024llamav32} and \textbf{\Gemma} \cite{Gemma_Team2025-ga} were selected as decoder-only dense IT models, while \textbf{\TFiveGemma} \cite{Zhang2025-ya} serves as a representative of recent encoder-decoder IT models. Since most of these LLMs are predominantly English- and/or Chinese-centric, we included \textbf{\Aya} \cite{Dang2024-nk}, which was pre-trained on extensive multilingual data, and \textbf{\Tower} \cite{Rei2025-ij}, a translation-specific LLM fine-tuned on Qwen-2.5-72B. For baseline comparison, we selected two neural machine translation models, \textbf{\NllbSmall} and \textbf{\NllbBig} \cite{NLLB_Team2022-cc}, along with a widely used proprietary system, \textbf{\GoogleTranslate}\footnote{Available at \url{https://translate.google.com/}. We consider \GoogleTranslate as a translation LLM since Google claims it is supported by LLMs \cite{Caswell22024}.}. 

\paragraph{Evaluation Metrics} Considering their popularity, we used COMET-22 \cite{rei-etal-2022-comet} and SacreBLEU \cite{post-2018-call} as the main evaluation metrics for our LLM translation outputs. chrF++ scores \cite{popovic-2017-chrf} were included in Appendix \ref{appendix_fig_table} as references for morphologically-rich target languages.

\subsection{Inference Details} \label{sec:inf_details}

We used vLLM \cite{Kwonetal2023} for inference with most models, with the exception of DeepSeek-V3.2-Exp, \TFiveGemma, and the baseline systems. For these models, we obtained inference results using their respective APIs or the HuggingFace Transformers library \cite{wolf-etal-2020-transformers}. We initially conducted experiments using Prompt 0 with the temperature and top\_p both set to 1. We further evaluated the effect of varying the temperature by increasing it to 1.5 and decreasing it to 0. Increasing the temperature to 1.5 resulted in a clear performance degradation across all language pairs, as measured by both COMET and BLEU scores. Conversely, setting the temperature to 0 led to slight performance improvements for nearly all language pairs. Consequently, all reported experiments were conducted with a temperature of 0. In the few-shot setting, we randomly selected 5 examples for each language pair from the rest of the corpora as demonstrations inserted in the prompt templates. 

With the exception of DeepSeek-V3.2-Exp and \GoogleTranslate, all models were run without quantization on 4 NVIDIA GH200 GPUs. On average, an IT model requires approximately 10 minutes to process one LP (3,000 instances), whereas a reasoning model requires about 18 minutes.

\section{Evaluation Results} \label{results}

\begin{figure*}[ht]
  \centering
  \includegraphics[width=0.9\textwidth]{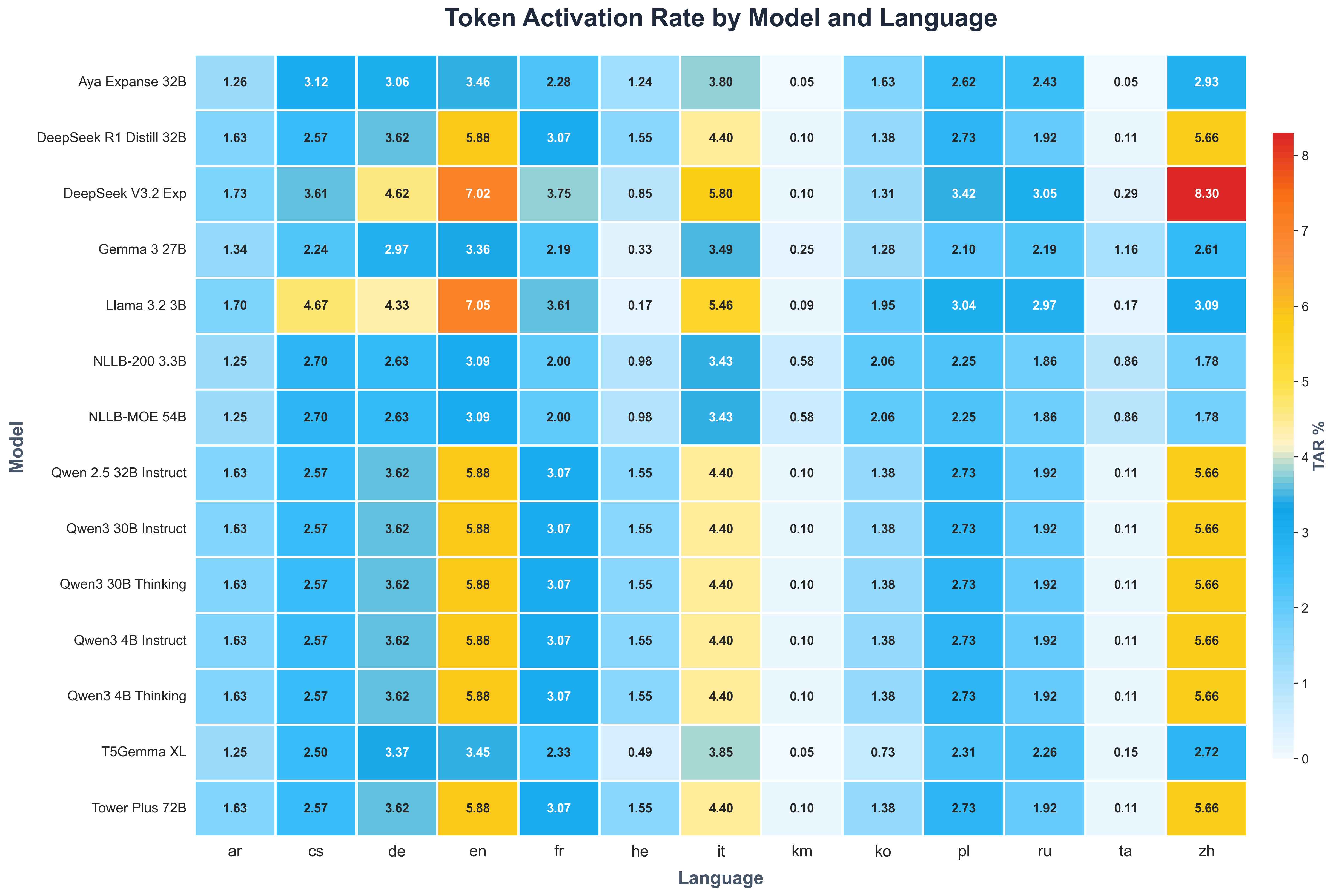}
  \caption{TAR for 13 different languages and 14 models (excluding \GoogleTranslate).}
  \label{fig.token_coverage}
\end{figure*}

This section presents the results of our evaluation. Figure \ref{fig.zero-shot_comet_p2} displays COMET scores for all 22 LPs under the zero-shot setting.

The parallel coordinates plot in Figure \ref{fig.zero-shot_comet_p2} reveals interesting patterns in COMET scores across language pairs of varying resource availability and across LLMs trained for general versus translation-specific purposes. Detailed tables of COMET and BLEU scores for both zero-shot and few-shot settings exhibit consistent patterns and are therefore provided in Appendix \ref{appendix_fig_table}.

First, we observe that non-English-centric LPs have substantially lower average COMET scores than English-centric pairs, with greater performance variability across these LPs. This reflects the current state of the art in MT, namely the English-centricity of language resources. The figure also shows clear performance degradation for most LLMs on LPs involving lower-resource languages, such as Arabic-Hebrew, English-Tamil, and Khmer-English, suggesting that resource availability plays a key role in translation performance. However, we also observe that certain LPs, such as Chinese-French, yield notably lower COMET scores than French-Italian, despite both involving high-resource languages. We hypothesize that typological distance also influences COMET scores. In Section \ref{analysis}, we further investigate whether language resource availability, using TAR as a proxy, and typological distance are significant factors of LLM performance in translation.

Regarding model-wise performance, translation-specific LLMs such as \Tower and \GoogleTranslate achieve the highest COMET scores for most LPs, generally outperforming general-purpose LLMs. Among general-purpose models, those that are large in scale and trained on multilingual data such as \Aya, \Gemma, and \DeepseekChat, achieve results comparable to translation-specific LLMs. This further suggests that greater exposure to diverse language data during training may positively impact translation performance, a hypothesis we explore in the following section.

\section{Analysis and Findings} \label{analysis}

This section investigates factors associated with LLM failure in translation, especially for low-resource languages. The previous section suggests that factors such as language resource availability and typological distance between languages may be important predictors of LLM translation performance. We explore these factors in Sections \ref{coverage} and \ref{typo_distance}. Assuming that language data representation in the training data is an important factor for LLM performance, we further investigate whether generating more tokens (\textit{i.e.}, the number of reasoning tokens) at test time can compensate for limited TAR during pre-training in Section \ref{compensation}.

\begin{table*}[]
\centering
\resizebox{13.5cm}{!}{%
\begin{tabular}{ccccccccc}
\toprule
Model                                     & TAR & GENETIC & GEOGRAPHIC       & SYNTACTIC        & PHONOLOGICAL & INVENTORY & FEATURAL         & MEAN    \\ \hline
\QwenLargeInstruct         & \textbf{0.5352} & -0.1294 & -0.2395          & -0.2605          & 0.1940       & 0.0982    & -0.4134          & -0.1736 \\
\QwenLargeThink         & \textbf{0.5339} & -0.1032 & -0.2402          & -0.2453          & 0.2225       & 0.1010    & \textbf{-0.4275} & -0.1599 \\
\QwenSmallInstruct              & \textbf{0.6575} & -0.1302 & -0.0915          & \textbf{-0.4974} & 0.1583       & 0.0615    & \textbf{-0.4723} & -0.1470 \\
\QwenSmallThink             & \textbf{0.6490}  & -0.1196 & -0.1687          & -0.4127          & 0.2140       & 0.0866    & \textbf{-0.4963} & -0.1594 \\
\Llama         & \textbf{0.7206} & -0.0682 & -0.1286          & -0.4216          & 0.2668       & -0.0539   & \textbf{-0.5666} & -0.1486 \\
\Gemma                    & \textbf{0.5164} & -0.1706 & -0.3282          & -0.1792          & 0.1478       & 0.1586    & -0.3691          & -0.2157 \\
\QwenOld                         & \textbf{0.6693}  & -0.0761 & -0.0799         & \textbf{-0.4937}          & 0.1830        & -0.0148    & \textbf{-0.4720} & -0.1353 \\
\DeepseekQwen & \textbf{0.6685} & -0.1932 & -0.1026          & \textbf{-0.5949} & 0.0548       & 0.0666    & \textbf{-0.4635} & -0.2038 \\
\Aya               & \textbf{0.5545} & -0.2598 & -0.3132          & -0.2977          & 0.0355       & 0.1484    & -0.3759          & -0.2746 \\
\Tower                   & \textbf{0.5954} & -0.2347 & -0.2158          & \textbf{-0.4974} & 0.0117       & 0.1164    & \textbf{-0.4281} & -0.2593 \\
\TFiveGemma         & \textbf{0.5905} & -0.1857 & -0.0649          & \textbf{-0.5403} & 0.1237       & -0.0076   & \textbf{-0.4533} & -0.1707 \\
\DeepseekChat                             & 0.3166          & -0.1842 & -0.3243          & -0.1863          & 0.1240       & 0.1495    & -0.3528          & -0.2234 \\
\DeepseekReasoner                         & \textbf{0.4700}   & -0.0191 & -0.2189          & -0.2002          & 0.2706       & 0.0397    & \textbf{-0.4292} & -0.1224 \\
\NllbSmall                   & \textbf{0.5643} & -0.2850 & \textbf{-0.5233} & -0.2289          & -0.0040       & 0.0968    & \textbf{-0.4307} & -0.4101 \\
\NllbBig                    & \textbf{0.5080}  & -0.2621 & \textbf{-0.5037} & -0.2168          & -0.0328      & 0.1037    & -0.4011          & -0.3950 \\
\bottomrule
\end{tabular}%
}
\caption{Pearson's $r$ correlation between COMET scores and TAR, genetic, geographic, syntactic, phonological, inventory, featural and the mean of the latter six typological distances. \textbf{Bold values} are statistically significant.}
\label{tab:comet_vocab_typo}
\end{table*}

\subsection{Token Activation Rate} \label{coverage}

Since we do not know the actual distribution of each language in the training data, we leveraged our test data as samples to calculate the Token Activation Rate (TAR) of the model vocabulary as an approximation, to understand language resource availability during training. TAR measures the proportion of a model's tokenizer vocabulary that is activated when processing text in a given language. Formally, given a model $M$ with vocabulary $V_M$, a tokenizer function $\mathrm{Tokenize}_M$, and text data $D_l$ in language $l$, TAR is defined as:

\begin{equation} \label{eq:tar}
    \mathrm{TAR}(l, M) = \frac{|\{t \in V_M : t \in \mathrm{Tokenize}_M(D_l)\}|}{|V_M|}
\end{equation}

We used the 3,000 instances per language pair from either the source or the target in the test set, and tokenized them into input IDs using the corresponding model tokenizers. We retained only unique input IDs for each language (13 in total) and divided this count by the vocabulary size of the model. For example, we used the source text of the 3,000 instances in Arabic-Hebrew, tokenizing them with the \QwenSmallInstruct tokenizer to obtain 2,469 unique input IDs. This count was then divided by the model vocabulary size of 151,669, resulting in a TAR of 1.63\% for Arabic.

Figure \ref{fig.token_coverage} presents a heatmap of TAR across the 13 languages and 14 models. It reveals that Khmer, Tamil, and Hebrew exhibit notably low TAR across nearly all models, which corresponds precisely to the COMET score drops observed for Arabic-Hebrew, English-Tamil, and Khmer-English in Figure \ref{fig.zero-shot_comet_p2}. Regarding model-wise coverage, neural MT models such as NLLB maintain better balance across languages compared to English- and Chinese-dominant LLMs, resulting in smaller performance disparities among LPs.

\begin{figure*}[ht]
  \centering
  \includegraphics[width=0.8\textwidth]{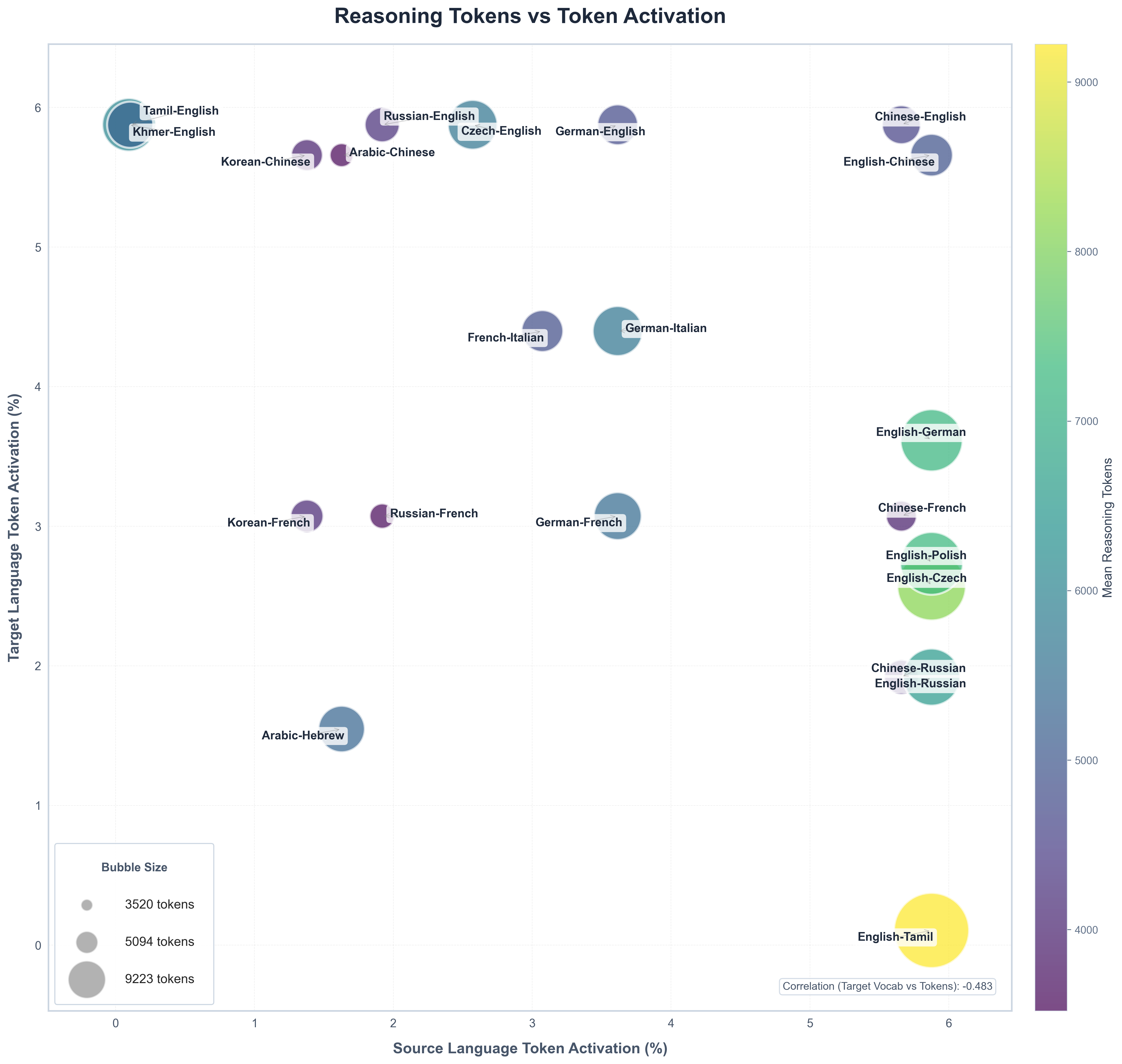}
  \caption{TAR of the vocabulary of \QwenSmallThink per language pair in the source (X axis) and target (Y axis) language against the average number of reasoning tokens.}
  \label{fig.n_token_vs_coverage}
\end{figure*}

\subsection{Typological Distance} \label{typo_distance}

We observe that although Chinese, French, and Italian exhibit high TAR, the average COMET scores for Chinese-French are lower than those for French-Italian. We hypothesize that other factors, such as typological distance, also affect LLM performance. To quantify these distances across LPs, we rely on URIEL \cite{littell-etal-2017-uriel}, a database and toolkit that provides multiple distance measures between languages, including genetic, geographic, syntactic, phonological, inventory, and featural distances. These measures capture, respectively, genealogical relatedness within a language family, physical distance between speaker populations, divergence in grammatical structure, differences in sound systems, variation in phoneme inventories, and an overall typological distance derived from the full set of URIEL features. Details of the design and computation of the distances can be found in Littell et al \shortcite{littell-etal-2017-uriel}.

Table \ref{tab:comet_vocab_typo} displays Pearson's $r$ correlation scores between COMET scores, TAR\footnote{TAR for a language pair is computed by summing the TAR values of the source and target languages.}, the six typological distances and their mean. With the exception of \DeepseekChat, TAR is highly correlated with COMET scores across all models. Syntactic and featural distances also exhibit moderate negative correlations with model performance for many models. That means, greater distance between two languages corresponds to lower COMET scores. The correlation patterns for BLEU and chrF++ scores are consistent with these observations, as shown in Tables \ref{tab:bleu_vocab_typo} and \ref{tab:chrf_vocab_typo} in Appendix \ref{app_BLEU_vs_token_cov}. These results align with prior findings reported by Khiu et al. \shortcite{khiu-etal-2024-predicting}, Ploeger et al. \shortcite{ploeger-etal-2025-cross}, and Hirak et al. \shortcite{hirak-etal-2026-assessing}. 

\subsection{Reasoning Tokens} \label{compensation}

Given that low TAR in a model's vocabulary at the pre-training stage is highly correlated with translation performance, we analyze whether reasoning LLMs would generate more reasoning tokens for languages with lower TAR as a compensatory mechanism. Furthermore, we also explore whether generating more reasoning tokens at test time would improve translation quality.

\begin{figure*}[ht]
  \centering
  \includegraphics[width=0.8\textwidth]{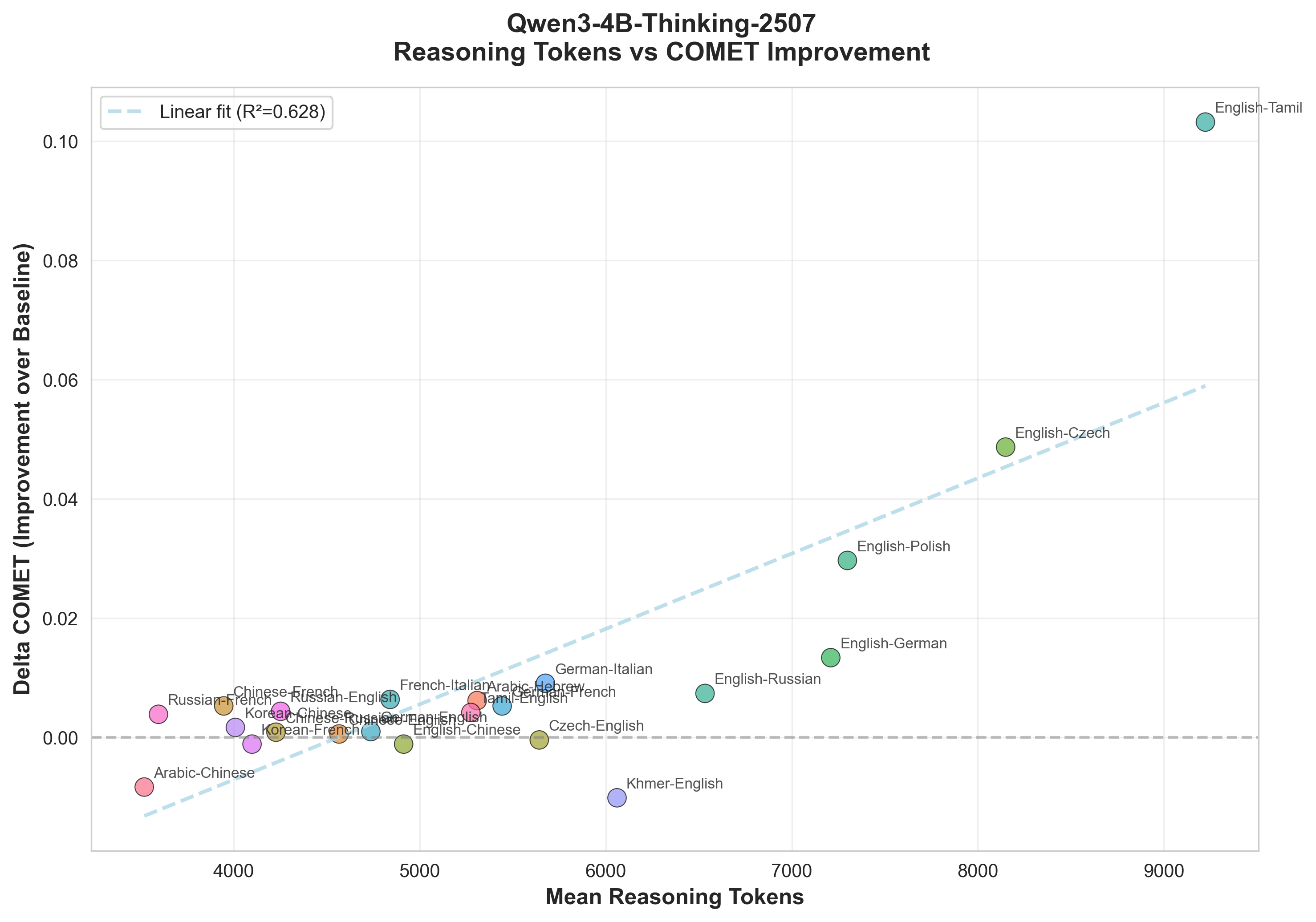}
  \caption{The average number of reasoning tokens from \QwenSmallThink \textit{vs} the increase of COMET scores ($\Delta$COMET) compared to its IT model \QwenSmallInstruct.}
  \label{fig.n_token_vs_delta_comet}
\end{figure*}

\paragraph{Reasoning Tokens \textit{vs} TAR} Figure \ref{fig.n_token_vs_coverage} illustrates the relationship between TAR for each LP and the average number of reasoning tokens generated by \QwenSmallThink, with source language TAR on the X-axis and target language TAR on the Y-axis. The figure clearly shows that \QwenSmallThink\ generates substantially fewer reasoning tokens for LPs with high TAR on the target side, such as Korean-Chinese and Russian-English. For LPs with high source-side TAR but medium or low target-side TAR, at the mid-right region of the figure, the model generates considerably more reasoning tokens. We further calculated correlations between the number of reasoning tokens and TAR on both the source and target sides for the 4 reasoning models. We find that TAR in the target language is indeed negatively correlated with the number of reasoning tokens ($r$=-0.2572, $\rho$=-0.3177, $\tau$=-0.2306; all statistically significant). This indicates that lower TAR in the target language tends to elicit more reasoning tokens at test time as compensation. 

\paragraph{Reasoning Tokens \textit{vs} Metric Improvements} We continued our investigation on whether more reasoning tokens generated at test time would benefit the performance of LLM translation, by examining the difference of COMET and BLEU scores ($\Delta$COMET and $\Delta$BLEU) between reasoning models and their instruction-tuned counterparts. This analysis examines whether increases or decreases in COMET and BLEU scores correlate with the number of generated reasoning tokens. 

\begin{table}[H]
\centering
\resizebox{6cm}{!}{%
\begin{tabular}{ccc}
\toprule
Model Name                              & $\Delta$COMET & $\Delta$BLEU  \\ \hline
\QwenLargeThink         & \textbf{0.5734}  & 0.3273  \\
\QwenSmallThink              & \textbf{0.7925}  & \textbf{0.5900}  \\
\DeepseekQwen & -0.1043 & 0.0177  \\
\DeepseekChat                        & \textbf{-0.9825} & \textbf{-0.9660} \\
\bottomrule
\end{tabular}%
}
\caption{Pearson's $r$ correlation between $\Delta$COMET and $\Delta$BLEU and the average number of reasoning tokens for each LP. \textbf{Bold values} are statistically significant.}
\label{tab:corr_delta_mean}
\end{table}

Table \ref{tab:corr_delta_mean} presents Pearson's $r$ correlation coefficients between the average number of reasoning tokens and $\Delta$COMET and $\Delta$BLEU. The table reveals that their correlations are model-dependent. For Qwen models, more reasoning tokens exhibit a strong positive correlation with COMET score improvements, indicating that additional reasoning tokens contribute positively to translation quality. Figure \ref{fig.n_token_vs_delta_comet} plots the relationship between $\Delta$COMET and the average number of reasoning tokens for \QwenSmallThink, showing that a simple linear model could explain 62.8\% of the variability in the response variable. For language pairs with low TAR at the target side like English-Tamil, the model generates a considerable amount of reasoning tokens, which correlates positively with the increase of COMET scores. However, DeepSeek models, in contrast, exhibit negative correlations. To further explore this model-specific difference, we continued our investigations in Section \ref{sec:ablative_reasoning} on other reasoning models.

\section{Validation on Token Activation Rate}

The analyses in Section \ref{coverage} rely on the assumption that TAR reflects how well a language is represented in the model's pre-training data. To validate this assumption, we sought open-source LLMs that disclose language-level data distributions. To our best effort, we identified Bloomz \cite{BigScience_Workshop2022-zv} and EuroLLM \cite{Martins2024-xc}, both of which report this information. Other open-source LLMs including Olmo \cite{groeneveld-etal-2024-olmo} and Apertus \cite{Apertus2025-xc} do not explicitly provide detailed language distributions in their training data.

\begin{table}[ht]
\centering
\resizebox{4cm}{!}{%
\begin{tabular}{ccc}
\toprule
Language   & Actual  & TAR    \\ \hline
Arabic     & 4.65\%  & 2.58\% \\
English    & 30.11\% & 3.63\% \\
French     & 12.94\% & 2.56\% \\
Chinese    & 16.21\% & 4.17\% \\
Tamil      & 0.50\%  & 1.73\% \\
Gujarati   & 0.07\%  & 2.30\% \\
Hindi      & 1.53\%  & 2.88\% \\
Malayalam  & 0.23\%  & 2.46\% \\
Portuguese & 4.92\%  & 4.78\% \\
Telugu     & 0.19\%  & 2.34\% \\
\bottomrule
\end{tabular}%
}
\caption{TAR and the actual language-level training data distribution (Actual) in bloomz-7b1.}
\label{tab:bloomz_dis}
\end{table}

\begin{table}[ht]
\centering
\resizebox{4cm}{!}{%
\begin{tabular}{ccc}
\toprule
Language & Actual  & TAR    \\ \hline
German   & 6.00\%  & 6.06\% \\
French   & 6.00\%  & 4.23\% \\
Italian  & 6.00\%  & 7.28\% \\
Chinese  & 3.50\%  & 3.88\% \\
Russian  & 2.50\%  & 4.32\% \\
Polish   & 2.50\%  & 5.34\% \\
Arabic   & 1.50\%  & 1.87\% \\
Korean   & 1.50\%  & 2.27\% \\
Czech    & 1.50\%  & 4.99\% \\
English  & 82.50\% & 6.52\% \\
\bottomrule
\end{tabular}%
}
\caption{TAR and the actual language-level training data distribution (Actual) in EuroLLM-22B-Instruct-2512.}
\label{tab:eurollm_dis}
\end{table}

As shown in Table \ref{tab:bloomz_dis} for bloomz-7b1, we computed TAR for Arabic, English, French, Chinese, and Tamil using the method and data described in Sections \ref{coverage} and \ref{sec:data} respectively. To increase the number of languages for validation, we incorporated additional language data including Gujarati, Hindi, Malayalam, Portuguese and Telugu from the monolingual training data of WMT24 \cite{kocmi-etal-2024-findings}, as these are mostly from similar sources and of comparable length to our data. For EuroLLM-22B-Instruct-2512, the training data distributions for German, French, Italian, Chinese, Russian, Polish, Arabic, Korean, Czech and English are openly released. We computed their TAR using our data and present the results in Table \ref{tab:eurollm_dis}. 

We then applied a leave-one-language-out methodology: for each language, we remove it from the set and recompute the correlation between TAR and actual training data proportions. This tests whether the observed correlation is robust or driven by individual outlier languages.

\begin{table}[ht]
\centering
\resizebox{5cm}{!}{%
\begin{tabular}{cccc}
\toprule
left-out  & $r$        &  $\rho$      & $\tau$     \\ \hline
\textit{None}       & 0.4980          & \textbf{0.7697} & \textbf{0.5556} \\
Arabic     & 0.4925          & \textbf{0.7500}     & \textbf{0.5556} \\
English    & 0.5215          & \textbf{0.7500}     & \textbf{0.5556} \\
French     & 0.5444           & \textbf{0.8167} & \textbf{0.6111} \\
Chinese    & 0.4166          & \textbf{0.7500}     & \textbf{0.5556} \\
Tamil      & 0.4514          & \textbf{0.7833} & \textbf{0.6111} \\
Gujarati   & 0.4661          & \textbf{0.7333} & 0.5000         \\
Hindi      & 0.5036          & \textbf{0.7500}     & \textbf{0.5556} \\
Malayalam  & 0.4761          & \textbf{0.7333} & 0.5000         \\
Portuguese & \textbf{0.7544} & \textbf{0.8167} & \textbf{0.6111} \\
Telugu     & 0.4688          & \textbf{0.7333} & 0.5000  \\  
\bottomrule
\end{tabular}%
}
\caption{Pearson's $r$, Spearman's $\rho$ and Kendall's $\tau$ correlation coefficients between the actual language-level training data distribution and TAR of bloomz-7b1. Leave-one-language-out was applied to ensure the score stability. \textbf{Bold values} are statistically significant. }
\label{tab:bloomz_corr}
\end{table}

\begin{table}[ht]
\centering
\resizebox{5cm}{!}{%
\begin{tabular}{cccc}
\toprule
left-out & $r$ & $\rho$        & $\tau$         \\ \hline
\textit{None}      & 0.4177  & \textbf{0.6669} & \textbf{0.5320} \\
German    & 0.4581  & 0.5899          & 0.4490          \\
French    & 0.4138  & \textbf{0.7866} & \textbf{0.6286} \\
Italian   & 0.5389  & 0.6156          & 0.5089          \\
Chinese   & 0.4077  & \textbf{0.7105} & \textbf{0.5880} \\
Russian   & 0.4130  & \textbf{0.7246} & \textbf{0.6086} \\
Polish    & 0.4417  & \textbf{0.6901} & 0.5477          \\
Arabic    & 0.4159  & 0.5814          & 0.4490          \\
Korean    & 0.4050  & 0.5814          & 0.4490          \\
Czech     & 0.4314  & \textbf{0.7695} & \textbf{0.6286} \\
English   & 0.6581  & 0.5719          & 0.4642          \\
\bottomrule
\end{tabular}%
}
\caption{Pearson's $r$, Spearman's $\rho$ and Kendall's $\tau$ correlation coefficients between the actual language-level training data distribution and TAR of EuroLLM-22B-Instruct-2512. Leave-one-language-out was applied to ensure the score stability. \textbf{Bold values} are statistically significant.}
\label{tab:eurollm_corr}
\end{table}

Tables \ref{tab:bloomz_corr} and \ref{tab:eurollm_corr} display the Pearson's $r$, Spearman's $\rho$ and Kendall's $\tau$ correlation coefficients between the actual training data distribution and TAR, for bloomz-7b1 and EuroLLM-22B-Instruct-2512. The Spearman and Kendall rank correlations are consistently strong and statistically significant across most leave-one-language-out conditions for both models, indicating that the relationship is robust and not driven by individual outlier languages. The Pearson correlations are generally weaker, which is expected given the non-linear relationship between TAR and actual data proportions (e.g., English has a disproportionately high data share but its TAR is bounded). These results support using TAR as a reliable proxy for language representation in the training data, though we note the limitation that our validation is restricted to only two models with 10 languages each.

\section{Validation on Reasoning Tokens} \label{sec:ablative_reasoning}

To validate the generality of our findings on Qwen and DeepSeek models regarding the relationship between TAR, the number of reasoning tokens, and $\Delta$COMET and $\Delta$BLEU, we replicated our analysis on two additional reasoning LLMs, Olmo-3-7B-Think and K2-Think-V2, along with their instruction-tuned counterparts, Olmo-3-7B-Instruct and K2-V2-Instruct \cite{Olmo2025-cf,K2_Team2026-sp}.

\paragraph{Reasoning Tokens \textit{vs} TAR} We observe consistent negative correlations between the TAR of the target language and the average number of reasoning tokens ($r = -0.3045$, $\rho = -0.4917$, $\tau = -0.3414$), all statistically significant. These results corroborate our earlier findings: reasoning LLMs tend to generate more tokens when translating into languages with lower token activation rates. This suggests that increased reasoning token usage may act as a compensatory mechanism for limited token availability on the target side.

\begin{table*}[]
\centering
\resizebox{9cm}{!}{%
\begin{tabular}{ccccc}
\toprule
Model    & Metric        & $r$       & $\rho$    & $\tau$ \\ \hline
\multirow{2}{*}{K2-Think-V2}     & $\Delta$COMET & 0.0698           & 0.3755           & 0.2814       \\
                                 & $\Delta$BLEU  & \textbf{-0.4367} & \textbf{-0.4241} & -0.2814      \\
\multirow{2}{*}{Olmo-3-7B-Think} & $\Delta$COMET & -0.0376          & -0.0271          & -0.0087      \\
                                 & $\Delta$BLEU  & -0.0100          & -0.0717          & -0.0736   \\
\bottomrule
\end{tabular}%
}
\caption{Pearson's $r$, Spearman's $\rho$ and Kendall's $\tau$ correlation scores between $\Delta$COMET, $\Delta$BLEU and the average number of reasoning tokens for K2-Think-V2 and Olmo-3-7B-Think. \textbf{Bold values} are statistically significant.}
\label{tab:reasoning_ablation}
\end{table*}

\paragraph{Reasoning Tokens \textit{vs} Metric Improvements} Table \ref{tab:reasoning_ablation} reports the Pearson, Spearman, and Kendall correlations between $\Delta$COMET, $\Delta$BLEU, and the average number of reasoning tokens for K2-Think-V2 and Olmo-3-7B-Think. Consistent with our observations on Qwen and DeepSeek models, the relationship between the number of reasoning tokens and translation quality measured by COMET and BLEU is highly model-dependent. For some models (e.g., \QwenSmallThink), increased reasoning tokens are associated with improvements in COMET and BLEU scores, whereas for others (e.g., \DeepseekReasoner and K2-Think-V2), the correlations are weak or negative.

This variability is expected, as translation performance of LLMs depends on multiple factors, including training data, model architecture, and alignment strategies \textit{etc}. Furthermore, automatic metrics such as COMET and BLEU are sensitive to output noise. As observed in models like \Gemma and K2-V2-Instruct, the inclusion of explanatory text alongside translations (see Appendix \ref{appendix_noise}) can distort metric scores and obscure the true relationship between reasoning and translation quality. These findings highlight the importance of careful model selection and output cleaning to ensure valid evaluation and reliable conclusions. Overall, our results suggest that while increased reasoning token usage consistently compensates for low TAR, its impact on translation quality is not universal, underscoring the need to jointly consider token dynamics and model-specific factors when evaluating reasoning LLMs for MT. 

\section{Conclusion}

In this paper, we systematically evaluated the performance of LLMs on MT, with a focus on understanding their failures in low-resource and non-English-centric settings. To better characterize language representation within model vocabularies, we introduced TAR and validated it as a proxy using models with known training language distributions. Our analyses show that TAR and typological distance are both strongly associated with translation quality: lower TAR and greater typological distance consistently correlate with reduced COMET and BLEU scores. We further examined the relationship between TAR, the number of reasoning tokens, and translation quality. Our results indicate that increased reasoning token generation is closely associated with low TAR in the target language, suggesting a compensatory mechanism. However, the extent to which additional reasoning tokens improve COMET and BLEU scores is highly model-dependent, highlighting the influence of other factors such as training data, alignment, and output noise. Overall, our findings emphasize the importance of token-level dynamics in understanding multilingual performance in LLMs. For future work, we plan to develop robust methods for controlling output noise and to investigate additional factors affecting multilingual capabilities, particularly from an interpretability perspective.

\section*{Limitations}

Despite our findings, several limitations should be noted. First, output noise remains a significant challenge. LLM-generated translations often include extraneous text, and the extent of such noise varies across models and prompting strategies. Although we design prompts and apply rule-based filtering to encourage translation-only outputs, we cannot guarantee complete removal of noise. As a result, automatic evaluation metrics such as COMET and BLEU may be affected, potentially introducing bias into our results. Second, while we show that TAR correlates with known language distributions and translation performance, it does not fully capture all aspects of multilingual competence. Therefore, TAR should be interpreted as a complementary signal rather than a complete explanation of model behavior. Third, metrics such as COMET and BLEU, while widely used, are sensitive to surface variation and may not fully capture semantic adequacy, especially in multilingual and low-resource settings. This limitation is further exacerbated by the presence of output noise and multiple valid translations. 

Finally, our study focuses on correlation rather than causation. While we identify strong relationships between TAR, reasoning token usage, and translation performance, we do not establish causal mechanisms. Future work is needed to develop controlled experiments and model interventions to better understand the causal role of token dynamics in multilingual generation.

\section*{Sustainability Statement}

Following the principles of ``Green AI'' \cite{Schwartz2020}, we aim to minimize the environmental impact of our experiments by improving inference efficiency. Specifically, we leverage vLLM to accelerate inference and reduce computational overhead. In total, our experiments require approximately 200 GPU hours, corresponding to an energy consumption of 397.64 kWh and an estimated 3.03 kg of CO$_2$ emissions, calculated using the methodology of Lannelongue et al \shortcite{Lannelongue2021}.

\section*{Acknowledgments}

This work has received funding from the European Union's Horizon Europe research and innovation programme under the Marie Skłodowska-Curie grant agreement No. 101126636. 

The computations were performed on resources provided through Sigma2 – the national research infrastructure provider for high-performance computing and large-scale data storage in Norway. We acknowledge Norway and Sigma2 for awarding this project access to the Olivia supercomputer, through Project nn9851k.


\bibliography{eamt26,anthology-1,anthology-2}
\bibliographystyle{eamt26}

\clearpage 
\onecolumn 
\appendix
\counterwithin{figure}{section}
\setcounter{figure}{0} 
\counterwithin{table}{section}
\setcounter{table}{0} 

\section{Appendix: Dataset Details} \label{appendix_data}

\begin{table}[ht]
\centering
\resizebox{10cm}{!}{%
\begin{tabular}{ccc}
\toprule
Lang\_pairs             & Test\_size & Source                            \\ \hline
Arabic-Chinese (ar-zh)  & 3,000      & TED Multilingual Parallel Corpora \\
Arabic-Hebrew (ar-he)   & 3,000      & TED Multilingual Parallel Corpora \\
Chinese-French (zh-fr)  & 3,000      & TED Multilingual Parallel Corpora \\
Chinese-Russian (zh-ru) & 3,000      & TED Multilingual Parallel Corpora \\
French-Italian (fr-it)  & 3,000      & SwissAdmin                        \\
German-French (de-fr)   & 3,000      & SwissAdmin                        \\
German-Italian (de-it)  & 3,000      & SwissAdmin                        \\
Korean-Chinese (ko-zh)  & 3,000      & Chinese-Korean Parallel Corpus    \\
Korean-French (ko-fr)   & 3,000      & TED Multilingual Parallel Corpora \\
Russian-French (ru-fr)  & 3,000      & TED Multilingual Parallel Corpora \\
English-Chinese (en-zh) & 3,000      & WMT20 QE Shared Task              \\
English-Czech (en-cs)   & 3,000      & WMT20 QE Shared Task              \\
English-German (en-de)  & 3,000      & WMT20 QE Shared Task              \\
English-Polish (en-pl)  & 3,000      & WMT20 QE Shared Task              \\
English-Russian (en-ru) & 3,000      & WMT20 QE Shared Task              \\
English-Tamil (en-ta)   & 3,000      & WMT20 QE Shared Task              \\
Chinese-English (zh-en) & 3,000      & WMT20 QE Shared Task              \\
Czech-English (cs-en)   & 3,000      & WMT20 QE Shared Task              \\
German-English (de-en)  & 3,000      & WMT20 QE Shared Task              \\
Khmer-English (km-en)   & 3,000      & WMT20 QE Shared Task              \\
Russian-English (ru-en) & 3,000      & WMT20 QE Shared Task              \\
Tamil-English (ta-en)   & 3,000      & WMT20 QE Shared Task              \\ \bottomrule
\end{tabular}%
}
\caption{The size of our test set for each language pair and their corresponding sources.}
\label{tab:data_details}
\end{table}

\section{Appendix: LLM Output Noise}  \label{appendix_noise}

\begin{figure}[ht]
\centering
\fbox{%
  \begin{minipage}{13cm}
  \textbf{English Source}: WTI crude futures were last down by 0.9\% to \$55.91 after falling as low as \$54.79. Brent crude futures dropped 1.4\% to \$61.89. \\
  \textbf{Czech Translation}: Závěry futures na ropu WTI klesly o 0,9 \% na 55,91 dolarů poté, co dosáhly minima 54,79 dolarů. Futures na ropu Brent klesly o 1,4 \% na 61,89 dolarů.\textbackslash n\textbackslash n**Explanation of choices:**\textbackslash n\textbackslash n* **Závěry futures:**  \"Futures\" translates well to \"futures\" in Czech, but to be clearer we specify it's the *conclusions* or *settlements* of the futures contracts.\textbackslash n* **klesly o:**  \"were down by\" translates naturally to \"klesly o\" (fell by).\textbackslash n* **poté, co dosáhly minima:** \"after falling as low as\" is best expressed as \"after reaching a minimum.\"\textbackslash n* **dolarů:**  \"dollars\" is \"dolarů\" in Czech.
  \end{minipage}
  }
  \caption{Noise in LLM Output from \Gemma using Prompt 0.}
\label{MT_noise}
\end{figure}

\section{Appendix: Prompt Templates} \label{appendix_prompt}

\textbf{Prompt 0}

\{src\_lang\}: \{src\_txt\}

\{tgt\_lang\}:

\noindent \textbf{Prompt 1}

Translate the following \{src\_lang\} into \{tgt\_lang\}: \{src\_text\}

\noindent \textbf{Prompt 2}

Translate the following \{src\_lang\} into \{tgt\_lang\} and only output the target text: \{src\_text\}

\section{Appendix: LLM Output Noise Detection} \label{appendix_verifier}

We introduce a rule-based method to calculate the proportion of instances that contain only the translation without extra explanatory text or text in an incorrect language, to quantitatively detect noise in LLM outputs. We term this metric the clean translation rate (Clean\%). MT outputs containing extra explanatory text were detected using regular expressions matching explanatory terms such as ``explanation'', ``indicate'', and ``analysis''. Outputs in the wrong target language were identified based on a language identification model from fastText \cite{bojanowski-etal-2017-enriching} with a confidence threshold of 60\%. An instance is classified as a clean translation only when it contains neither extra explanatory text nor text in an incorrect target language with more than 60\% confidence. The clean translation rate is formally defined in Equation \ref{eq:clean}:

\begin{equation} \label{eq:clean}
    \mathrm{Clean}\% = \frac{N-|E \cup W|}{N}
\end{equation}

\noindent where is $N$ is the total number of instances, $E$ is the set of instances containing explanatory text and $W$ is the set containing text in the wrong language. $\mathrm{Exp}\%$ and $\mathrm{WrongL}\%$ are defined as $\frac{|E|}{N}$ and $\frac{|W|}{N}$.

\begin{table}[h]
\centering
\resizebox{15cm}{!}{%
\begin{tabular}{cccc|ccc|ccc}
\toprule
                                         & \multicolumn{3}{c}{Clean\% $\uparrow$}                            & \multicolumn{3}{c}{Expl\% $\downarrow$}                          & \multicolumn{3}{c}{WrongL\% $\downarrow$}                  \\
Model\_name                              & Prompt 0         & Prompt 1         & Prompt 2         & Prompt 0        & Prompt 1        & Prompt 2        & Prompt 0 & Prompt 1        & Prompt 2         \\ \hline
\QwenLargeInstruct         & 95.98\%          & 96.69\%          & \textbf{96.72\%} & 2.95\%          & 2.69\%          & \underline{\textbf{2.62\%}} & 1.18\%   & \textbf{0.64\%} & 0.68\%           \\
\QwenLargeThink         & 96.70\%          & \textbf{96.74\%} & 96.72\%          & 2.81\%          & \textbf{2.76\%} & 2.78\%          & 0.65\%   & \textbf{0.51\%} & 0.52\%           \\
\QwenSmallInstruct              & 95.35\%          & \textbf{96.23\%} & 96.21\%          & \textbf{3.03\%} & 3.09\%          & 3.08\%          & 1.77\%   & \textbf{0.70\%} & 0.77\%           \\
\QwenSmallThink              & \textbf{96.49\%} & 96.34\%          & 96.38\%          & \textbf{2.84\%} & 2.94\%          & 2.91\%          & 0.99\%   & 0.75\%          & \textbf{0.73\%}  \\
\Llama         & 59.42\%          & 30.46\%          & \textbf{92.91\%} & 29.47\%         & 67.60\%         & \textbf{3.30\%} & 21.67\%  & 15.87\%         & \textbf{4.03\%}  \\
\Gemma                   & 2.68\%           & 0.27\%           & \textbf{96.76\%} & 97.32\%         & 99.72\%         & \textbf{2.79\%} & 31.82\%  & 31.82\%         & \textbf{0.46\%}  \\
\QwenOld                & 44.38\%          & 71.95\%          & \textbf{90.94\%} & 53.93\%         & 26.71\%         & \textbf{7.21\%} & 15.22\%  & 8.08\%          & \textbf{4.10\%}  \\
\DeepseekQwen & 73.25\%          & 84.10\%          & \textbf{95.81\%} & 22.22\%         & 14.42\%         & \textbf{2.89\%} & 11.92\%  & 5.18\%          & \textbf{1.36\%}  \\
\Aya               & 83.37\%          & 68.29\%          & \textbf{96.35\%} & 15.03\%         & 27.46\%         & \textbf{3.10\%} & 4.93\%   & 12.47\%         & \textbf{0.60\%}  \\
\Tower                   & 94.85\%          & 94.74\%          & \textbf{96.12\%} & 3.34\%          & 4.58\%          & \textbf{3.33\%} & 2.05\%   & 0.94\%          & \textbf{0.58\%}  \\
\TFiveGemma         & 53.68\%          & 72.36\%          & \textbf{82.65\%} & 33.88\%         & 12.86\%         & \textbf{4.11\%} & 25.41\%  & 19.35\%         & \textbf{14.23\%} \\
\DeepseekChat             & 39.28\%          & /                & \underline{\textbf{96.81\%}} & 59.17\%         & /               & \textbf{2.86\%} & 11.65\%  & /               & \underline{\textbf{0.37\%}}  \\
\DeepseekReasoner         & /                & /                & 95.41\%          & /               & /               & 4.29\%          & /        & /               & 1.90\%    \\ 
\bottomrule
\end{tabular}%
}
\caption{The clean translation rate (Clean\%), the rate of generating extra explanatory texts (Expl\%) and the rate of outputting wrong language (WrongL\%) for different prompts and LLMs. We did not run all prompts on DeepSeek-V3.2-Exp as we see much better performance on other LLMs using Prompt 2.}
\label{tab:clean_rate}
\end{table}

Table \ref{tab:clean_rate} shows Clean\% across different prompts and models. The results demonstrate that DeepSeek-V3.2-Exp exhibits the strongest performance in translation instruction following, and Prompt 2 yields the cleanest translation output among the three prompt templates. This finding was confirmed by manual inspection, and Prompt 2 was therefore used for all subsequent experiments and analyses. 

\section{Appendix: Additional Evaluation Results} \label{appendix_fig_table}

\begin{table*}[h]
\centering
\resizebox{16cm}{!}{%
\begin{tabular}{cccccccccccc|ccccccc|ccccccc}
\toprule
& \multicolumn{11}{c}{non-English centric} & \multicolumn{7}{c}{En-XX} & \multicolumn{7}{c}{XX-En} \\
model\_name                      & ar-he  & ar-zh  & de-fr  & de-it  & fr-it  & ko-fr  & ko-zh  & ru-fr  & zh-fr  & zh-ru  & mean   & en-cs  & en-de  & en-pl  & en-ru  & en-ta  & en-zh  & mean   & cs-en  & de-en  & km-en  & ru-en  & ta-en  & zh-en  & mean   \\ \hline
\QwenLargeInstruct      & 0.7278 & 0.7440  & 0.8202 & 0.8524 & 0.8615 & 0.6684 & 0.8902 & 0.7152 & 0.7153 & 0.7401 & 0.7735 & 0.8707 & 0.8621 & 0.8710 & 0.8795 & 0.8554 & 0.8828 & 0.8703 & 0.8306 & 0.8680  & 0.7966 & 0.8589 & 0.8570 & 0.8323 & 0.8406 \\
\QwenLargeThink      & 0.7108 & 0.7352 & 0.8193 & 0.8528 & 0.8622 & 0.6636 & 0.8907 & 0.7163 & 0.7153 & 0.7401 & 0.7706 & 0.8824 & 0.8650 & 0.8791 & 0.8821 & 0.8729 & 0.8808 & 0.8771 & 0.8266 & 0.8668 & 0.7876 & 0.8602 & 0.8577 & 0.8309 & 0.8383 \\
\QwenSmallInstruct           & 0.6630 & 0.7321 & 0.8026 & 0.8341 & 0.8506 & 0.6555 & 0.8806 & 0.7048 & 0.7052 & 0.7316 & 0.7560 & 0.7839 & 0.8379 & 0.8140 & 0.8594 & 0.6724 & 0.8785 & 0.8077 & 0.8145 & 0.8611 & 0.7354 & 0.8502 & 0.8311 & 0.8278 & 0.8200 \\
\QwenSmallThink           & 0.6692 & 0.7238 & 0.8079 & 0.8432 & 0.8570  & 0.6544 & 0.8823 & 0.7087 & 0.7105 & 0.7325 & 0.7589 & 0.8326 & 0.8513 & 0.8437 & 0.8668 & 0.7756 & 0.8774 & 0.8412 & 0.8141 & 0.8621 & 0.7253 & 0.8546 & 0.8353 & 0.8284 & 0.8200   \\
\Llama            & 0.5145 & 0.6595 & 0.7528 & 0.7840  & 0.8263 & 0.6111 & 0.7953 & 0.6639 & 0.6698 & 0.6372 & 0.6914 & 0.6763 & 0.7986 & 0.7786 & 0.7432 & 0.7199 & 0.7954 & 0.7520 & 0.7914 & 0.8446 & 0.6399 & 0.8298 & 0.7982 & 0.7948 & 0.7831 \\
\Gemma                   & 0.7771 & 0.7363 & 0.8259 & 0.8596 & 0.8655 & 0.6713 & 0.8894 & 0.7208 & 0.7175 & 0.7423 & 0.7806 & 0.9021 & 0.8686 & 0.8989 & 0.8912 & 0.9010 & 0.8793 & 0.8902 & 0.8348 & 0.8659 & 0.8129 & 0.8602 & 0.8641 & 0.8309 & 0.8448 \\
\bottomrule
\end{tabular}%
}
\caption{\textbf{COMET} scores of translations for 22 language pairs using Prompt 2 under \textbf{zero-shot setting}.}
\label{tab:zero-shot_COMET_P2}
\end{table*}

\begin{table}[h]
\centering
\resizebox{16cm}{!}{%
\begin{tabular}{cccccccccccc|ccccccc|ccccccc}
\toprule
                                         & \multicolumn{11}{c}{non-English centric}                                              & \multicolumn{7}{c}{En-XX}                             & \multicolumn{7}{c}{XX-En}                             \\ 
model\_name                              & ar-he & ar-zh & de-fr & de-it & fr-it & ko-fr & ko-zh & ru-fr & zh-fr & zh-ru & mean  & en-cs & en-de & en-pl & en-ru & en-ta & en-zh & mean  & cs-en & de-en & km-en & ru-en & ta-en & zh-en & mean  \\ \hline
\QwenLargeInstruct         & 10.34 & 17.72 & 28.87 & 26.46 & 26.69 & 11.31 & 32.95 & 20.09 & 14.50 & 7.50  & 19.64 & 25.39 & 32.34 & 21.96 & 24.34 & 5.50  & 39.88 & 24.9  & 28.27 & 37.60 & 17.20 & 37.79 & 19.89 & 28.02 & 28.13 \\
\QwenLargeThink         & 9.30  & 18.28 & 29.18 & 26.02 & 27.18 & 11.55 & 33.12 & 19.40 & 14.87 & 7.19  & 19.61 & 26.19 & 33.94 & 21.58 & 24.91 & 6.77  & 39.02 & 25.40 & 28.67 & 38.96 & 17.41 & 38.65 & 21.40 & 28.96 & 29.01 \\
\QwenSmallInstruct              & 4.44  & 15.68 & 24.52 & 20.88 & 23.94 & 10.05 & 30.08 & 17.67 & 12.28 & 6.62  & 16.62 & 15.86 & 27.15 & 12.84 & 20.73 & 0.65  & 37.83 & 19.18 & 25.26 & 36.07 & 9.70  & 35.09 & 14.99 & 26.99 & 24.68 \\
\QwenSmallThink              & 6.44  & 16.93 & 25.89 & 22.88 & 25.33 & 10.92 & 31.19 & 18.42 & 13.75 & 6.30   & 17.81 & 19.10  & 29.58 & 17.22 & 20.90 & 2.86  & 37.64 & 21.22 & 26.31 & 36.60 & 10.88 & 36.57 & 17.41 & 27.31 & 25.85 \\
\Llama         & 0.50  & 3.62  & 18.12 & 12.90 & 22.35 & 3.41  & 9.96  & 13.60 & 8.15  & 2.45  & 9.51  & 12.03 & 24.00    & 13.77 & 12.04 & 2.12  & 23.09 & 14.51 & 25.05 & 35.47 & 3.16  & 33.36 & 11.73 & 21.71 & 21.75 \\
\Gemma                    & 13.45 & 18.48 & 31.38 & 28.88 & 28.40 & 11.83 & 34.61 & 21.45 & 14.28 & 7.24  & 21.00    & 32.31 & 36.23 & 26.59 & 27.24 & 9.71  & 41.07 & 28.86 & 30.49 & 40.98 & 20.49 & 39.48 & 23.78 & 29.67 & 30.81 \\
\bottomrule
\end{tabular}%
}
\caption{\textbf{BLEU} scores of translations for 22 language pairs using Prompt 2 under \textbf{zero-shot setting}.}
\label{tab:zero-shot_BLEU_P2}
\end{table}

\begin{table}[h]
\centering
\resizebox{16cm}{!}{%
\begin{tabular}{cccccccccccc|ccccccc|ccccccc}
\toprule
                                         & \multicolumn{11}{c}{non-English centric}                                              & \multicolumn{7}{c}{En-XX}                             & \multicolumn{7}{c}{XX-En}                             \\ 
model\_name                              & ar-he & ar-zh & de-fr & de-it & fr-it & ko-fr & ko-zh & ru-fr & zh-fr & zh-ru & mean  & en-cs & en-de & en-pl & en-ru & en-ta & en-zh & mean  & cs-en & de-en & km-en & ru-en & ta-en & zh-en & mean  \\ \hline
\QwenLargeInstruct         & 30.32                              & 12.85                              & 52.50                              & 52.44                              & 52.21                              & 30.94                              & 22.11                              & 38.98                              & 35.88                              & 27.98                              & 35.62                             & 50.24                              & 57.58                              & 48.62                              & 49.27                              & 39.27                              & 26.28                              & 45.21                             & 55.08                              & 62.63                              & 43.52                              & 62.27                              & 50.71                              & 55.62                              & 54.97                             \\
\bottomrule
\end{tabular}%
}
\caption{\textbf{chrF++} scores of translations for 22 language pairs using Prompt 2 under \textbf{zero-shot setting}.}
\label{tab:zero-shot_chrf_P2}
\end{table}

\begin{table}[h]
\centering
\resizebox{16cm}{!}{%
\begin{tabular}{cccccccccccc|ccccccc|ccccccc}
\toprule
                                         & \multicolumn{11}{c}{Non-English centric}                                                                                                                                                            & \multicolumn{7}{c}{En-XX}                                                                                                   & \multicolumn{7}{c}{XX-En}                                                                                                   \\
model\_name                              & ar-he           & ar-zh           & de-fr           & de-it           & fr-it           & ko-fr           & ko-zh           & ru-fr           & zh-fr           & zh-ru           & mean            & en-cs           & en-de           & en-pl           & en-ru           & en-ta           & en-zh           & mean            & cs-en           & de-en           & km-en           & ru-en           & ta-en           & zh-en           & mean            \\ \hline
\QwenLargeInstruct        & 0.7193          & 0.7410          & 0.8207          & 0.8393          & 0.8559          & 0.6667          & 0.8900          & 0.7018          & 0.7105          & 0.7369          & 0.7682          & 0.8146          & 0.8434          & 0.8702          & 0.8749          & 0.8457          & 0.8499          & 0.8498          & 0.8109          & 0.8595          & 0.7921          & 0.8511          & 0.8563          & 0.8316          & 0.8336          \\
\QwenLargeThink         & 0.7180          & 0.7358          & 0.8203          & 0.8538          & 0.8629          & 0.6640          & 0.8899          & 0.7122          & 0.7154          & 0.7420          & 0.7714          & 0.8832          & 0.8635          & 0.8765          & 0.8836          & 0.8737          & 0.8700          & 0.8751          & 0.8270          & 0.8661          & 0.7902          & 0.8590          & 0.8596          & 0.8323          & 0.8390          \\
\bottomrule
\end{tabular}%
}
\caption{\textbf{COMET} scores of translations for 22 language pairs using Prompt 2 under \textbf{few-shot setting}. The three baseline models do not have few-shot settings and they are here for comparison purpose.}
\label{tab:few-shot_COMET_P2}
\end{table}

\begin{table}[h]
\centering
\resizebox{16cm}{!}{%
\begin{tabular}{cccccccccccc|ccccccc|ccccccc}
\toprule
\multicolumn{1}{l}{}                     & \multicolumn{11}{c}{non-English centric}                                                                                                                                                & \multicolumn{7}{c}{En-XX}                                                                                            & \multicolumn{7}{c}{XX-En}                                                                                            \\
model\_name                              & ar-he          & ar-zh          & de-fr          & de-it          & fr-it          & ko-fr          & ko-zh          & ru-fr          & zh-fr          & zh-ru         & mean           & en-cs          & en-de          & en-pl          & en-ru          & en-ta          & en-zh          & mean           & cs-en          & de-en          & km-en          & ru-en          & ta-en          & zh-en          & mean           \\ \hline
\QwenLargeInstruct         & 4.79           & 17.55          & 28.68          & 18.02          & 23.14          & 11.40          & 33.50          & 19.40          & 13.78          & 6.90          & 17.71          & 11.34          & 22.29          & 18.46          & 18.75          & 2.04           & 12.54          & 14.24          & 27.28          & 35.09          & 16.50          & 32.63          & 20.92          & 28.31          & 26.79          \\
\QwenLargeThink         & 8.89           & 18.11          & 29.22          & 26.27          & 27.45          & 11.72          & 33.53          & 19.36          & 14.89          & 7.48          & 19.69          & 26.33          & 33.63          & 19.66          & 25.18          & 8.25           & 33.39          & 24.41          & 29.68          & 39.33          & 17.76          & 39.12          & 22.74          & 29.65          & 29.71          \\
\bottomrule
\end{tabular}%
}
\caption{\textbf{BLEU} scores of translations for 22 language pairs using Prompt 2 under \textbf{few-shot setting}. The three baseline models do not have few-shot settings and they are here for comparison purpose.}
\label{tab:few-shot_BLEU_P2}
\end{table}

\begin{table}[h]
\centering
\resizebox{16cm}{!}{%
\begin{tabular}{cccccccccccc|ccccccc|ccccccc}
\toprule
\multicolumn{1}{l}{} & \multicolumn{11}{c}{non-English centric}                                                                                                                                                 & \multicolumn{7}{c}{En-XX}                                                                                            & \multicolumn{7}{c}{XX-En}                                                                                            \\
model\_name                              & ar-he          & ar-zh          & de-fr          & de-it          & fr-it          & ko-fr          & ko-zh          & ru-fr          & zh-fr          & zh-ru          & mean           & en-cs          & en-de          & en-pl          & en-ru          & en-ta          & en-zh          & mean           & cs-en          & de-en          & km-en          & ru-en          & ta-en          & zh-en          & mean           \\ \hline
\QwenLargeInstruct         & 25.19          & 12.65          & 52.26          & 48.32          & 50.91          & 30.82          & 22.35          & 38.03          & 35.40          & 27.55          & 34.35          & 37.77          & 52.88          & 45.08          & 46.73          & 29.25          & 18.06          & 38.29          & 52.97          & 61.59          & 42.61          & 60.36          & 51.28          & 55.65          & 54.08          \\
\QwenLargeThink         & 29.81          & 12.52          & 52.82          & 52.56          & 52.82          & 30.91          & 23.01          & 38.81          & 36.35          & 28.46          & 35.81          & 52.14          & 58.59          & 48.00          & 49.81          & 43.44          & 24.99          & 46.16          & 55.77          & 63.52          & 44.51          & 62.89          & 52.63          & 56.65          & 55.99          \\
\bottomrule
\end{tabular}%
}
\caption{\textbf{chrF++} scores of translations for 22 language pairs using Prompt 2 under \textbf{few-shot setting}. The three baseline models do not have few-shot settings and they are here for comparison purpose.}
\label{tab:few-shot_chrf_P2}
\end{table}

\section{Appendix: Correlation Between BLEU, chrF++, TAR and Typological Distances} \label{app_BLEU_vs_token_cov} 

\begin{table}[ht]
\centering
\resizebox{15cm}{!}{%
\begin{tabular}{ccccccccc}
\toprule
Model                                     & TAR & GENETIC         & GEOGRAPHIC       & SYNTACTIC        & PHONOLOGICAL & INVENTORY & FEATURAL         & MEAN             \\ \hline
\QwenLargeInstruct         & \textbf{0.6330}  & -0.2691         & -0.1451          & \textbf{-0.5860}  & 0.0095       & -0.0311   & \textbf{-0.4997} & -0.2780          \\
\QwenLargeThink         & \textbf{0.6406} & -0.2767         & -0.1485          & \textbf{-0.5781} & 0.0194       & -0.0550    & \textbf{-0.5101} & -0.2839          \\
\QwenSmallInstruct              & \textbf{0.6398} & -0.2202         & -0.0140          & \textbf{-0.6034} & 0.0374       & -0.0408   & \textbf{-0.4573} & -0.1883          \\
\QwenSmallThink              & \textbf{0.6405} & -0.2452         & -0.0671          & \textbf{-0.5962} & 0.0202       & -0.0530    & \textbf{-0.4781} & -0.2308          \\
\Llama         & \textbf{0.7496} & -0.3750          & -0.2964          & \textbf{-0.6180}  & 0.0052       & -0.0136   & \textbf{-0.6268} & -0.4013          \\
\Gemma                    & \textbf{0.6505} & -0.3172         & -0.2405          & \textbf{-0.5495} & -0.0040      & -0.0310    & \textbf{-0.5216} & -0.3419          \\
\QwenOld                & \textbf{0.5121} & -0.2194         & 0.0102           & -0.3912          & 0.0967       & -0.1340   & -0.2798          & -0.1301          \\
\bottomrule
\end{tabular}%
}
\caption{Pearson's $r$ correlation between BLEU scores and TAR, genetic, geographic, syntactic, phonological, inventory, featural and the mean of the latter six typological distances. \textbf{Bold values} are statistically significant.}
\label{tab:bleu_vocab_typo}
\end{table}

\begin{table}[ht]
\centering
\resizebox{15cm}{!}{%
\begin{tabular}{ccccccccc}
\toprule
Model                                     & TAR             & GENETIC          & GEOGRAPHIC       & SYNTACTIC        & PHONOLOGICAL & INVENTORY & FEATURAL         & MEAN             \\ \hline
\QwenLargeInstruct         & 0.3074          & -0.3853          & \textbf{-0.6543} & \textbf{-0.4471} & 0.0770        & 0.1005    & \textbf{-0.7294} & \textbf{-0.5461} \\
\QwenLargeThink        & 0.3033          & -0.3740           & \textbf{-0.6473} & \textbf{-0.4237} & 0.0975       & 0.0965    & \textbf{-0.7249} & \textbf{-0.5319} \\
\QwenSmallInstruct              & 0.3934          & -0.3851          & \textbf{-0.5475} & \textbf{-0.5740}  & 0.0595       & 0.0799    & \textbf{-0.7388} & \textbf{-0.5141} \\
\QwenSmallThink              & 0.3533          & -0.3787          & \textbf{-0.6066} & \textbf{-0.4995} & 0.0891       & 0.0909    & \textbf{-0.7488} & \textbf{-0.5267} \\
\Llama         & \textbf{0.7884} & -0.3469          & \textbf{-0.5676} & \textbf{-0.4961} & 0.0972       & 0.0102    & \textbf{-0.7484} & \textbf{-0.5110}  \\
\Gemma                    & \textbf{0.5798} & -0.4111          & \textbf{-0.6963} & -0.3872          & 0.0595       & 0.1233    & \textbf{-0.6990}  & \textbf{-0.5638} \\
\QwenOld                & \textbf{0.4320}  & -0.2736          & -0.4198          & \textbf{-0.4284} & 0.1984       & -0.1251   & \textbf{-0.6222} & -0.3906          \\
\bottomrule
\end{tabular}%
}
\caption{Pearson's $r$ correlation between chrF++ scores and TAR, genetic, geographic, syntactic, phonological, inventory, featural and the mean of the latter six typological distances. \textbf{Bold values} are statistically significant.}
\label{tab:chrf_vocab_typo}
\end{table}

\end{document}